\documentclass[lettersize,journal]{IEEEtran}
\usepackage{amsmath,amsfonts}
\usepackage{algorithmic}
\usepackage{algorithm}
\usepackage{array}
\usepackage{textcomp}
\usepackage{stfloats}
\usepackage{url}
\usepackage{verbatim}
\usepackage{graphicx}
\usepackage{cite}
\usepackage{subfigure}
\usepackage{float}
\usepackage{multirow}
\usepackage{multicol}
\usepackage[T1]{fontenc}
\usepackage{aecompl}
 \usepackage{hyperref} 
\usepackage {subcaption} 
\usepackage{threeparttable}
\usepackage[utf8]{inputenc}
\usepackage{tabularx}
\usepackage{soul, color, xcolor}
\soulregister{\cite}7 
\soulregister{\citep}7 
\soulregister{\citet}7 
\soulregister{\ref}7 
\soulregister{\pageref}7 

\begin{document}

\title{Large-Scale LiDAR-Inertial Dataset for Degradation-Robust High-Precision Mapping}

\author{
    Xiaofeng Jin\IEEEauthorrefmark{2}, \textit{Student Member, IEEE}, 
    Ningbo Bu\IEEEauthorrefmark{1}, 
    Jianfei Ge\IEEEauthorrefmark{1}, 
    Shijie Wang\IEEEauthorrefmark{1},
    Jiangjian Xiao\IEEEauthorrefmark{1}, 
    Matteo Matteucci\IEEEauthorrefmark{2},\textit{Member, IEEE}
    \thanks{
    \IEEEauthorblockN{
    Xiaofeng Jin\IEEEauthorrefmark{2}, 
    Ningbo Bu\IEEEauthorrefmark{1}, 
    Jianfei Ge\IEEEauthorrefmark{1}, 
    Shijie Wang\IEEEauthorrefmark{1},
    Jiangjian Xiao\IEEEauthorrefmark{1}, 
    Matteo Matteucci\IEEEauthorrefmark{2}
    }
    
    \IEEEauthorblockA{
    \IEEEauthorrefmark{1} Ningbo Institute of Materials Technology and Engineering (NIMTE), Chinese Academy of Sciences (CAS), China. \\
    \IEEEauthorrefmark{2} Politecnico di Milano, Italy. \\
    Emails: \{buningbo, gejianfei, wangshijie, xiaojj\}@Nimte.ac.cn, \{xiaofeng.jin, matteo.matteucci\}@polimi.it
    }
    }
    \thanks{Xiaofeng Jin and Ningbo Bu contributed equally to this work.}
    \thanks{Corresponding author: Jianfei Ge.}
    
}



\maketitle

\pagestyle{empty}  
\thispagestyle{empty} 

\begin{abstract}
LiDAR–Inertial Odometry (LIO) has demonstrated strong real-time performance and efficient mapping compared to traditional terrestrial laser scanners. Although recent advances driven by public datasets have improved LIO stability under specific degraded conditions, existing studies still lack systematic validation in real-world scenarios such as long-duration sequences, multiple degradation factors, and seamless indoor–outdoor transitions. This gap significantly limits their applicability to general purpose, high-precision mapping.

To bridge this gap, we present a large-scale high-precision LIO dataset collected in four diverse real-world environments that cover areas of 60,000-750,000 $m^2$. Using a custom backpack-mounted platform equipped with two multi-beam rotating LiDAR sensors, an industrial-grade IMU, and RTK-GNSS modules, we ensure precise spatiotemporal alignment through accurate calibration and hardware-level synchronization. The dataset encompasses a variety of environments, including structured buildings, tunnels, slopes, and urban scenes, to support seamless indoor-outdoor mapping. An average of 1,000 m and comprises over 15,000 frames per sequence, significantly outscaling existing benchmarks. We further propose a novel 6-DoF ground-truth generation pipeline by fusing SLAM-based optimization with RTK-GNSS anchoring. By leveraging oblique photogrammetry integrated with RTK-GNSS measurements,  we
obtain georeferenced reference maps at real-world scale. These maps are then used to align the LiDAR-based maps and to validate the low centimeter-level (<3 cm) accuracy of the generated trajectories. This dataset achieves breakthroughs in trajectory length, scene complexity, and ground-truth precision, providing a comprehensive benchmark for evaluating LIO systems
and improving their generalization in practical high-precision mapping scenarios. We will open access to the truth value generation code and dataset, which can be obtained from the following link: \url{https://github.com/CNITECH-CV-LAB/Backpack2025}
\end{abstract}


\begin{figure*}[t]
\centering
\includegraphics[width=18cm]{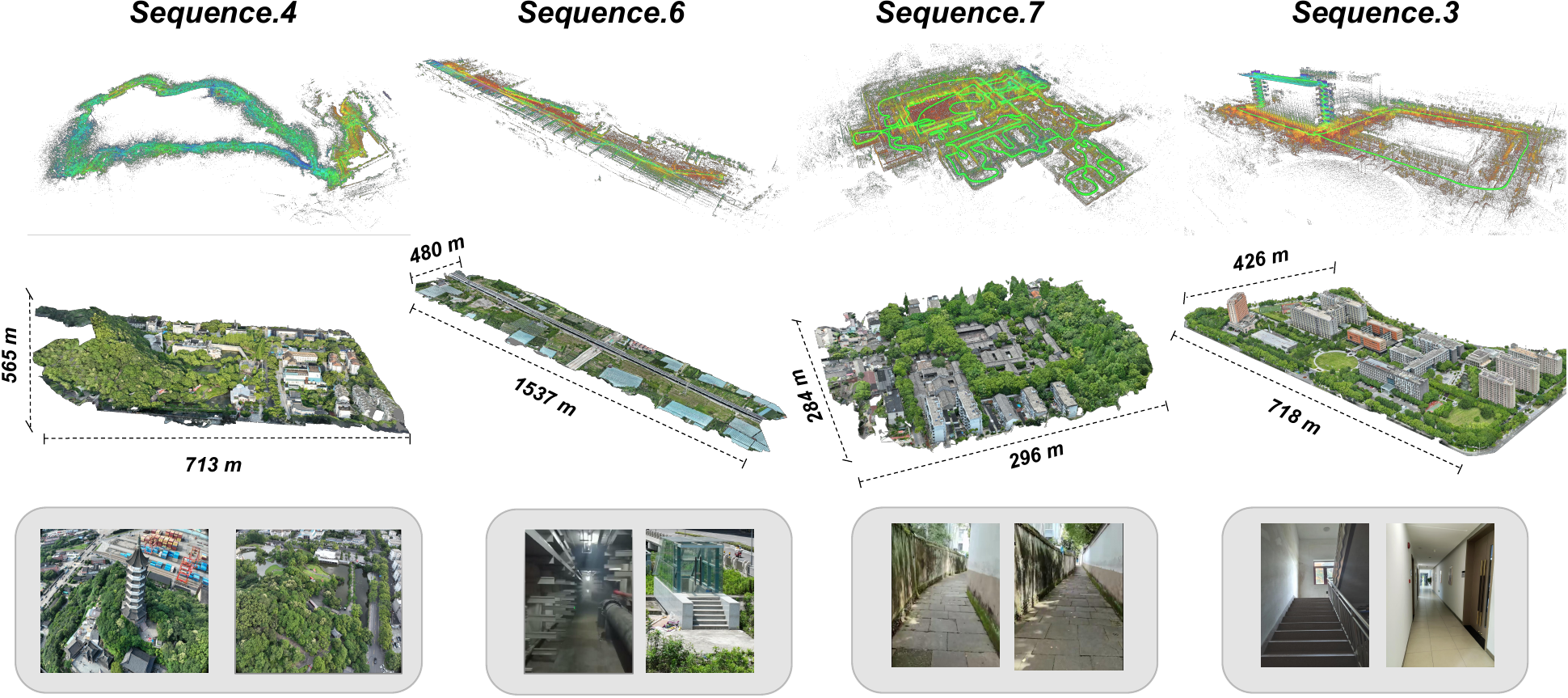}
\caption{Dataset Overview. The figure illustrates four large-scale and complex environments. The recorded sequences cover continuous scans across both indoor and outdoor areas. Challenging scenarios include dynamic objects on roads, narrow passages at staircases, and structures with similar scenes.}
\label{fig_1}
\end{figure*}

\section{Introduction}
\IEEEPARstart{A}{s} LiDAR technology evolves, Simultaneous Localization and Mapping (SLAM) systems that rely on LiDAR have gained significant importance in high-precision mapping applications. Compared with traditional stand-mounted laser scanners, LiDAR-Inertial Odometry (LIO) technology can greatly improve the efficiency and flexibility of mapping in large-scale and complex environments. As a result, LIO offers unique advantages for high-precision mapping applications such as urban digitization, infrastructure inspection, and post-disaster assessment. It is increasingly becoming an important alternative to traditional static mapping methods.

Several public LiDAR-IMU datasets released in recent years (e.g., Newer College\cite{NewerCollege}, Hilti SLAM\cite{HiltiSLAM}, 4Seasons\cite{Seasons}, Complex Urban\cite{ComplexUrban}, Montmorency\cite{Montmorency}, etc.) encompass typical degraded scenarios such as weak structural features, GNSS-denied environments, dynamic occlusions, and feature sparsity. These datasets serve as common benchmarks for extreme conditions, enabling continuous improvement of LIO algorithms and significantly enhancing their localization accuracy and robustness in such environments.

However, there is still a lack of real-world mapping datasets that comprehensively capture long-duration trajectories, diverse degraded conditions, and seamless transitions between indoor and outdoor environments. Although current algorithms perform well in isolated, controlled scenarios, their robustness over long sequences and across diverse scenes has not been systematically validated, largely due to the lack of datasets capturing complex, continuous environments. This gap impedes the advancement of LIO systems toward general-purpose, high-precision mapping applications.

A core challenge in constructing such benchmarks lies in obtaining accurate ground truth across large-scale, diverse environments. However, existing ground-truth acquisition methods have notable limitations. In small or static scenes (e.g., within a single building), researchers often employ high-precision terrestrial laser scanners or motion capture systems to record ground truth maps and trajectories. For instance, the Hilti SLAM\cite{HiltiSLAM} Challenge dataset uses uniform terrestrial laser scans at each site to provide millimeter-level accuracy. However, this approach does not scale well to expansive or complex environments. In contrast, outdoor driving datasets typically rely on differential GNSS/INS systems to deliver ground truth for vehicle trajectories. While such systems offer decimeter-level global positioning, they provide limited rotational accuracy and cannot operate in GNSS-denied environments such as indoors or urban canyons. In fact, many off-road or indoor–outdoor datasets lack any true 6-DoF ground truth—either providing only SLAM-based approximations or omitting trajectories altogether—making them unsuitable for rigorous odometry evaluation. This absence of dense, high-precision ground truth in complex, large-scale settings significantly limits the community’s ability to stress-test LIO systems under generalized and compounded challenges.

In this work, we present a new large-scale, high-precision LIO dataset (illustrated in Figure \ref{fig_1}) designed to support systematic benchmarking of LIO under real mapping conditions. Our dataset encompasses four extensive mapping sites, each including ultra-long continuous trajectories spanning both indoor and outdoor areas. Each sequence contains over 15,000 LiDAR frames and averages about 1.5 km in length, covering scenes from building interiors to complex terrain. We collected the data using a wearable multi-modal sensor platform that integrates a 3D LiDAR, RTK-GNSS, and IMU, enabling data acquisition across a wide range of terrains. Our main contribution is the proposal of a multi-stage ground truth generation pipeline that combines RTK-GNSS with LiDAR mapping, validated by aerial mapping, to obtain 6-DoF trajectories with centimeter-level accuracy (<3 cm) even indoors. The dataset, with dense, time-synchronized, high-precision trajectories and high-resolution point cloud maps, is one of the first datasets that can evaluate the performance of LIO algorithms across indoor and outdoor complex environments, and has significant research and engineering value. The main contributions of this work include:

\begin{itemize}
\item[$\bullet$] A large-scale LIO dataset covering long, continuous indoor–outdoor trajectories under challenging conditions such as sparse features, GNSS-denied areas, and multi-floor structures.
\end{itemize}
\begin{itemize}
\item[$\bullet$] A backpack-mounted sensor platform integrating multiple rotating LiDAR sensors, an IMU, and RTK-GNSS, enabling data capture in all-terrain scenarios beyond wheeled platforms.
\end{itemize}
\begin{itemize}
\item[$\bullet$] A centimeter-level ground-truth generation pipeline that combines RTK-GNSS with multi-stage LiDAR mapping, yielding dense 6-DoF trajectories even in GNSS-denied environments.
\end{itemize}
\begin{itemize}
\item[$\bullet$] A complete benchmarking suite of 8 sequences with synchronized LiDAR–IMU inputs and high-quality ground-truth trajectories and maps for real-world LIO evaluation.
\end{itemize}

\begin{table*}[htp]
\caption{Feature Comparison Between Public LiDAR–IMU Datasets and Our Dataset}
\label{table2}
\resizebox{18cm}{!}{
\begin{threeparttable}
\begin{tabular}{cccccccc}

\hline
\textbf{Dataset} & \textbf{Year} & \textbf{Platform}                                                       & \textbf{\begin{tabular}[c]{@{}c@{}}Scale \\ (km/seq)\end{tabular}} & \textbf{Environment}                                                          & \textbf{\begin{tabular}[c]{@{}c@{}}Length \\ (scans/seq)\end{tabular}} & \textbf{Degenerate Factors}                                                               & \textbf{Ground Truth}                                                      \\ \hline
Newer College\cite{NewerCollege}    & IROS (2020)   & Handheld                                                                & $\sim$0.45                                                         & \begin{tabular}[c]{@{}c@{}}Campus \\ (Indoor/Outdoor)\end{tabular}            & $\sim$4k                                                               & None                                                                                      & 6DoF ICP                                                                   \\
MulRan\cite{MulRan}           & ICRA (2020)   & Vehicle                                                                 & $\sim$6                                                            & Urban roads                                                                   & $\sim$12k                                                              & Sparse features                                                                            & GNSS SLAM                                                                  \\
M2DGR\cite{M2DGR}            & RAL (2021)    & UGV                                                                     & $\sim$0.3                                                          & \begin{tabular}[c]{@{}c@{}}Campus \\ (Indoor/Outdoor)\end{tabular}            & $\sim$3k                                                               & \begin{tabular}[c]{@{}c@{}}Constrained Fov;\\ Escalator\end{tabular}                      & \begin{tabular}[c]{@{}c@{}}Mocap/\\ Laser tracker/\\ RTK-GNSS\end{tabular} \\
Hilti-Oxford\cite{Hilti-Oxford}     & RAL (2022)    & Handheld                                                                & $\sim$0.2                                                          & \begin{tabular}[c]{@{}c@{}}Indoor hall;\\ Construction site\end{tabular}      & $\sim$2K                                                               & \begin{tabular}[c]{@{}c@{}}Constrained Fov;\\ Fast motion\end{tabular}                    & 6DoF ICP                                                                   \\
FusionPortable\cite{FusionPortable}   & IROS (2022)   & \begin{tabular}[c]{@{}c@{}}Quadruped Robot,\\ UGV,Handheld\end{tabular} & $\sim$0.6                                                          & \begin{tabular}[c]{@{}c@{}}Campus \\ (Indoor/Outdoor)\end{tabular}            & $\sim$4k                                                               & Sparse features                                                                            & \begin{tabular}[c]{@{}c@{}}Laser tracker/\\ RTK-GNSS\end{tabular}          \\
GRACO\cite{GRACO}            & RAL (2023)    & UGV,UAV                                                                 & $\sim$0.5                                                          & Campus (Outdoor)                                                              & $\sim$4k                                                               & Sparse features                                                                            & RTK-GNSS/INS                                                               \\
KITTI-360\cite{KITTI-360}        & TPAMI (2023)  & Vehicle                                                                 & 6–10                                                               & Urban roads                                                                   & $\sim$100k                                                             & Sparse features                                                                            & RTK-GNSS/INS                                                               \\
WHU-Helmet\cite{WHU-Helmet}       & TGRS (2023)   & Helmet                                                                  & 0.3-3                                                              & GNSS-denied sites                                                             & $\sim$15k                                                              & \begin{tabular}[c]{@{}c@{}}Fast motion;\\ Sparse features\end{tabular}                     & SLAM                                                                       \\
ENWIDE\cite{COIN-LIO}           & ICRA (2024)   & Handheld                                                                & $\sim$0.4                                                          & Flat field                                                                    & $\sim$3k                                                               & Sparse features                                                                            & 6DoF ICP                                                                   \\
GEODE\cite{GEODE}            & IJRR (2024)   & \begin{tabular}[c]{@{}c@{}}Handheld,\\ UGV,USV\end{tabular}             & $\sim$1                                                            & Indoor/Outdoor                                                                & $\sim$12k                                                              & \begin{tabular}[c]{@{}c@{}}Constrained Fov;\\ Sparse features\end{tabular}                 & \begin{tabular}[c]{@{}c@{}}6DoF ICP/\\ PPK-GNSS\end{tabular}               \\
BotanicGarden\cite{BotanicGarden}    & RAL (2024)    & UGV                                                                     & $\sim$0.5                                                          & Natural sites                                                                 & $\sim$5k                                                               & Unstructured features                                                                      & 6DoF ICP                                                                   \\ \hline
Ours             & 2025          & Backpack                                                                & $\sim$1.5                                                          & \begin{tabular}[c]{@{}c@{}}General sites \\ (Indoor\&outdoor)\end{tabular} & $\sim$15k                                                              & \begin{tabular}[c]{@{}c@{}}Sparse/Unstructured \\ features;\\ Constrained Fov\end{tabular} & \begin{tabular}[c]{@{}c@{}}LiDAR BA \\ \& \\ RTK-GNSS SLAM\end{tabular}   \\ \hline
\end{tabular}
 \begin{tablenotes}
        \footnotesize
        \item[*] Sparse features indicate weakly constrained structures in degraded scenarios (e.g., door frames in long corridors or poles in open fields) which are susceptible to loss under system parameters such as point downsampling
      \end{tablenotes}
    \end{threeparttable}}
\end{table*}

\section{RELATED WORK}
In high-precision mapping, robotic navigation, and related applications, LIO algorithms enable accurate localization and mapping construction in unknown environments by combining data from LiDAR and IMU. However, real-world mapping scenarios often present degraded conditions, such as sparse features, dynamic occlusions, and GPS-denied areas, that challenge system robustness. Ensuring reliable localization and maintaining global consistency under such complex conditions has become a central focus of LIO research. In response, numerous methods have been proposed from various perspectives, and several benchmark datasets have been developed to facilitate standardized evaluation.

For LiDAR-Inertial SLAM, the early LOAM\cite{LOAM} algorithm pioneered real-time mapping by using high-frequency scan matching, but its loosely coupled LiDAR-Inertial integration led to poor performance in texture-sparse or dynamic environments. Subsequent approaches tightly integrated IMU data to enhance system robustness and maintain global consistency. For example, LIO-SAM\cite{LIO-SAM} adopts a factor graph framework that combines IMU pre-integration with ICP-based loop closure detection, achieving high-precision, low-drift trajectories. BEV-LIO(LC)\cite{BEV-LIO} employs a point cloud projection approach similar to that used in IPAL\cite{IPAL}, converting LiDAR point clouds into bird’s-eye view (BEV) images and extracting local/global CNN features, which are fused into the iEKF and loop closure modules, significantly improving trajectory consistency and loop closure capability. Meanwhile, filter-based methods such as  FAST-LIO2\cite{FAST-LIO2} and Faster-LIO\cite{faster-LIO} utilize iterative Kalman filtering and direct scan-to-map matching to achieve real-time odometry at around 100 Hz, making them particularly suitable for high-speed motion or complex environments. 

However, under certain degraded conditions, these general pipelines may still fail, prompting the development of targeted solutions. For instance, AdaLIO\cite{adalio} dynamically adjusts voxel grid resolution and correspondence thresholds to prevent localization divergence in narrow corridors; DLIO\cite{DLIO} employs a continuous-time trajectory representation and coarse-to-fine point cloud matching strategy, making it suitable for resource-constrained platforms; COIN-LIO\cite{COIN-LIO} incorporates LiDAR intensity information to augment geometric features, improving localization in textureless or tunnel-like scenarios; NV-LIOM\cite{NV-LIOM} leverages point cloud normals and keyframe pose graph optimization to maintain robustness in repetitive indoor structures. For extremely degraded scenarios, methods like MM-LINS\cite{MM-LINS} and Voxel-SLAM\cite{Voxel-SLAM} propose a multi-submap mechanism: when significant drift occurs due to long-term features deprivation, a new submap is created to temporarily discard the drifting map, which is later merged through scene recognition to correct global drift. While each method excels in its targeted scenario, none yet offers a unified solution for complex environments involving GNSS denial, sparse features, repetitive structures, long-range navigation, and seamless indoor–outdoor transitions.

For LiDAR-Inertial SLAM Benchmarks, in recent years, the research community has released several representative LiDAR–Inertial datasets that serve as valuable resources for the validation and optimization of LIO algorithms. For example, KITTI-360\cite{KITTI-360} provides 6-DoF poses obtained from fused GNSS and IMU data on 73.7 km of suburban roads from the perspective of a vehicle. While highly valuable, it features sparse structures and infrequent loop closures, limiting its generalizability. Newer College\cite{NewerCollege} provides dense ground truth by using a handheld OS1-64 LiDAR together with a static BLK360 scanner in a mixed environment of arcades and outdoor gardens However, its sequences are relatively short—averaging less than 500 m or 7 minutes—making it unsuitable for long-trajectory testing. Other datasets, such as Hilti-Oxford\cite{Hilti-Oxford}, M2DGR\cite{M2DGR}, and Enwide\cite{COIN-LIO}, target specific scenarios like construction sites, indoor-outdoor transitions, or tunnels. Their ground truth is provided via terrestrial laser scanning, motion capture (MoCAP), or RTK systems, respectively. Yet many of these sequences remain short, and indoor versus outdoor data are often collected separately. Datasets like UrbanNav\cite{UrbanNav}, WHU-Helmet\cite{WHU-Helmet}, and GEODE\cite{GEODE} support dense ground truth and cover various degraded conditions, including urban canyons, forests, and tunnels, but typically focus on a single scenario. Notably, they lack long, continuous sequences involving indoor–outdoor transitions, and none provides a unified ground-truth trajectory covering all forms of composite degradation.

In general, these datasets have provided valuable support for studying LIO performance under specific degraded conditions. However, given the complexity of real-world mapping, current public datasets still share several common limitations. First, the sequence durations and traveled distances are often limited, making them less suitable for truly large-scale mapping tasks. Second, ground truth is typically obtained via static scanning or RTK/INS combinations, which struggle to provide both wide-area coverage and densely continuous 6-DoF trajectories. Third, long-duration, seamless indoor–outdoor transitions are missing, making it difficult to comprehensively validate the generalizability of LIO algorithms in truly complex environments.

To address these gaps, we present a novel dataset, specifically designed to serve as a standardized evaluation benchmark for LIO algorithms in general-purpose mapping scenarios. As summarized in Table \ref{table2}, our dataset is collected from four large-scale real-world sites, each covering an area of approximately 60,000 up to 750,000 $m^2$. These sites include diverse challenges such as dynamic traffic, unstructured features, sparse features in flat terrain, and limited field of view (FOV) in narrow space. These design choices give the dataset several key characteristics: long-duration
trajectories, continuous indoor–outdoor mapping, high-precision dense ground truth, and the concurrent presence of multiple degradation conditions. Overall, our dataset establishes a more comprehensive and challenging benchmark for evaluating the generalizability of LIO algorithms.

\begin{figure}[!t]
\centering
\subfigure[]{\includegraphics[width=8cm]{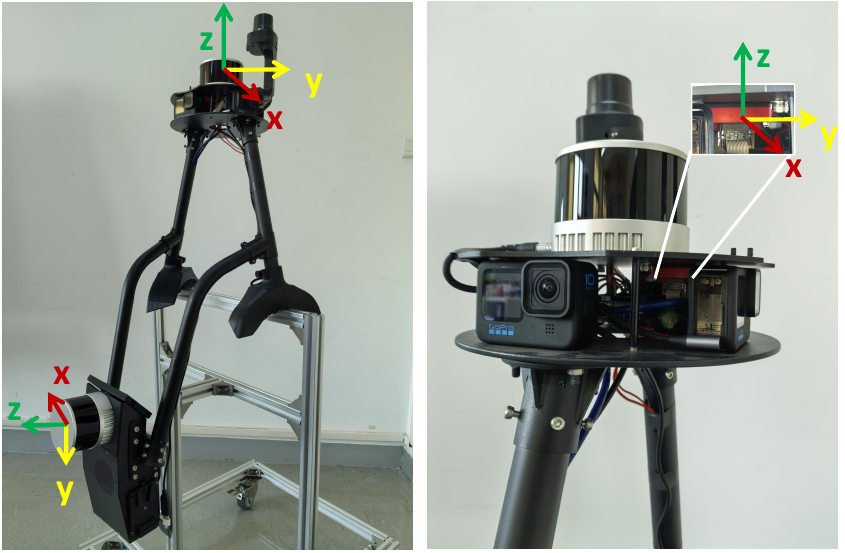}}
\subfigure[]{\includegraphics[width=8cm]{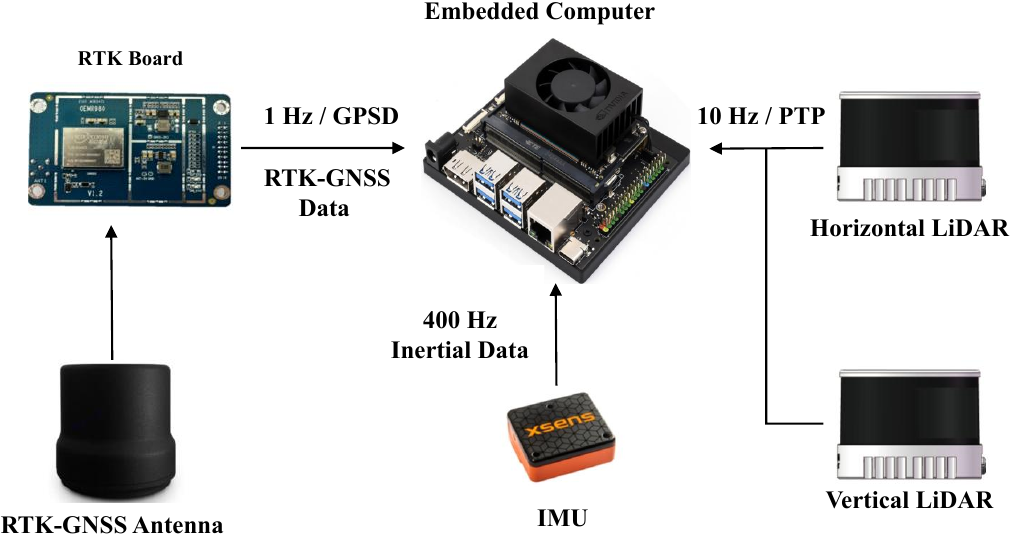}}
\caption{Data Acquisition Platform. (a) Sensor setup and mounting configuration of our system. (b) Clock synchronization architecture between the sensors.}
\label{fig_3}
\end{figure}

\section{METHODOLOGY}
To create a LIO-focused dataset with dense 6-DoF ground truth at low-centimeter accuracy in complex real-world scenarios, we built a complete data acquisition system — encompassing multi-modal sensor hardware design, spatiotemporal calibration, and ground-truth generation.

\subsection{Sensor Setup}
To ensure stable, continuous acquisition of high-precision multimodal data across complex indoor and outdoor terrains, we designed and implemented a backpack-type sensor platform. As summarized in Table \ref{hardware}, he platform integrates two LiDARs, an industrial-grade IMU, and RTK-GNSS modules, all fixed on a lightweight frame. As shown in Figure \ref{fig_3}(a), the LiDARs are mounted at the front and top of the rig to maximize the field of view. The IMU is positioned horizontally below the LiDARs, aligned with the direction of walking motion, to ensure stable readings. Meanwhile, the GNSS antenna is placed on top of the backpack to ensure optimal satellite reception in open areas. The sampling rates for the sensors are 10 Hz for each LiDAR, 400 Hz for the IMU, and 1 Hz for the RTK-GNSS. The overall weight of the platform is under 10 kg, allowing the operator to easily traverse various environments (e.g., staircases, narrow passages, forests) and ensuring that continuous indoor–outdoor trajectories can be recorded in full.

\subsection{Spatio-Temporal Calibration}
\label{calib}
To precisely align the multimodal data streams, we performed clock synchronization and extrinsic calibration for each sensor, ensuring temporal and spatial consistency for high- precision ground truth generation.

\subsubsection{Clock Synchronization}
The system uses the PPS (Pulse Per Second) signal from the GNSS receiver as the global time reference. As illustrated in Figure \ref{fig_3}(b), the GNSS receiver generates a PPS rising-edge pulse once per second which, combined with the UTC from NMEA messages, provides an absolute time reference for the host computer. The host uses the Chrony and GPSD service to discipline its Unix time to the PPS pulses, ensuring alignment with the GNSS UTC reference to within a few milliseconds.

For host–LiDAR synchronization, given a network environment with PTP (Precision Time Protocol) hardware support, we broadcast the host’s PPS-aligned time to the Hesai LiDARs using the ptp4l service. Each LiDAR’s clock source is set to PTP mode so that it accepts the PPS-synchronized network time. We also enable the LiDARs’ SyncAngle function, which aligns the 0° scan line of each rotation (the start of each frame) with the rising edge of the PPS pulse. This ensures the LiDARs start each scan cycle at an exact one-second boundary. In this way, both LiDARs share the same start-of-scan timing, achieving strict synchronization between the dual LiDARs.

For IMU time alignment, the host computer logs the IMU data via ROS, using the Unix time of arrival on the host as the IMU data timestamp. Since the IMU’s internal clock is not synchronized to the PPS, the host-recorded timestamp includes a fixed offset delay. We calibrate the time offset between LiDAR-IMU offline in the back-end through the method \cite{lidar-imu-init} and compensate for it in post-processing.

Through these steps, we establish a unified, PPS-driven time reference shared by the host, LiDARs, IMU, and GNSS, providing a solid temporal foundation for high-precision trajectory estimation.

\begin{table}[!t]
\caption{Hardware Specification of the Multi-Sensor System}
\label{hardware}
\centering
\begin{tabular}{cccccl}
\hline
\multicolumn{3}{c}{\textbf{Device}} & \multicolumn{3}{c}{\textbf{Type}} \\ \hline
     & Horizontal Lidar      &      &    & PandarXT-32              &   \\
     & Vertical LiDAR        &      &    & PandarXT-32              &   \\
     & IMU                   &      &    & Xsens MTi-630            &   \\
     & Embedded Computer     &      &    & Nvidia Jetson Xavier NX  &   \\
     & RTK-GNSS Antenna      &      &    & BT-468                 &   \\ \hline
\end{tabular}
\end{table}
\subsubsection{Extrinsics Calibration}

LiDAR–LiDAR extrinsic calibration is challenging when two sensors have little or no overlapping field of view due to their mounting arrangement. In such cases, a motion-based calibration approach is required. We adopt the DLBA-Calib framework \cite{lidar_cali}, which estimates the relative pose by minimizing point-to-plane residuals between one LiDAR’s scans and a global map
constructed by the other (reference) LiDAR:
\begin{equation}
\mathcal{L}_{\text{Global}} = \sum_{v \in \mathcal{V}} \omega_v \sum_{i=1}^{N_v} \left\| \mathbf{n}_v^\top \left( \mathbf{R}_{AB} \cdot \mathbf{p}^B_i + \mathbf{t}_{AB} - \mathbf{c}_v \right) \right\|^2
\end{equation}

\begin{figure}[!t]
\centering
\includegraphics[width=9cm]{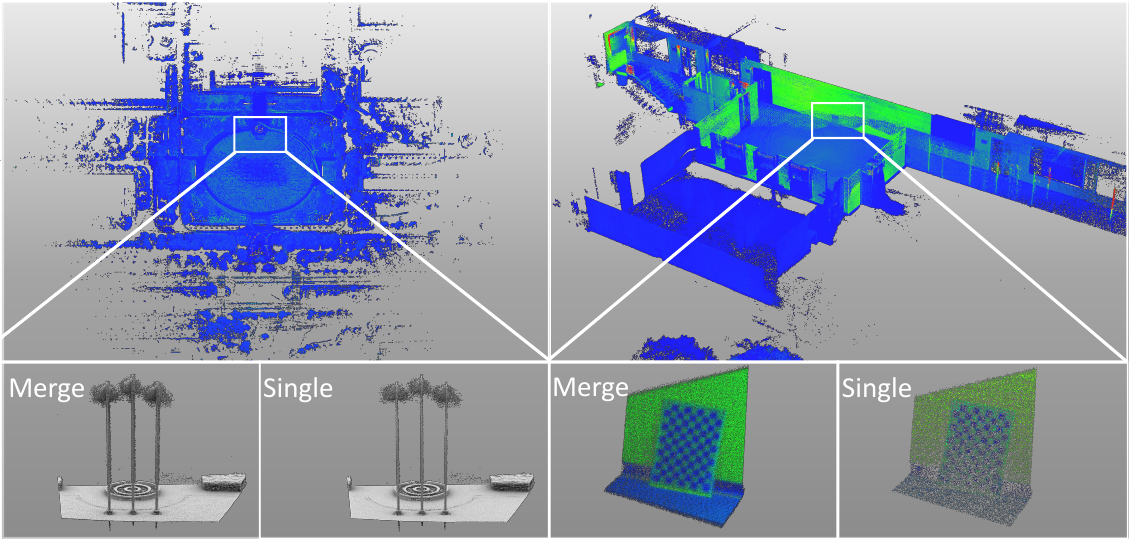}
\caption{Qualitative results of dual-LiDAR calibration. Maps are generated using dual- and single-LiDAR configurations in both indoor and outdoor environments. Reference landmarks are visualized to highlight the fusion errors.}
\label{lidar_cali}
\end{figure}

Here, $\mathbf{R}_{AB} \in \text{SO}(3)$ and $\mathbf{t}_{AB} \in \mathbb{R}^3$ denote the rotation and translation from LiDAR B to LiDAR A. $\mathbf{p}^B_i$ is a 3D point from LiDAR B. $(\mathbf{n}_v, \mathbf{c}_v)$ represent the normal vector and centroid of the $v$-th planar patch in the reference map; and $\omega_v$ is the confidence weight for that patch. The method tolerates initial misalignment up to 0.4 m and 30°, and as shown in Figure \ref{lidar_cali}, it can achieve calibration accuracy within 5 mm and 0.2°.

\begin{figure*}[htp]
\centering
\includegraphics[width=18cm]{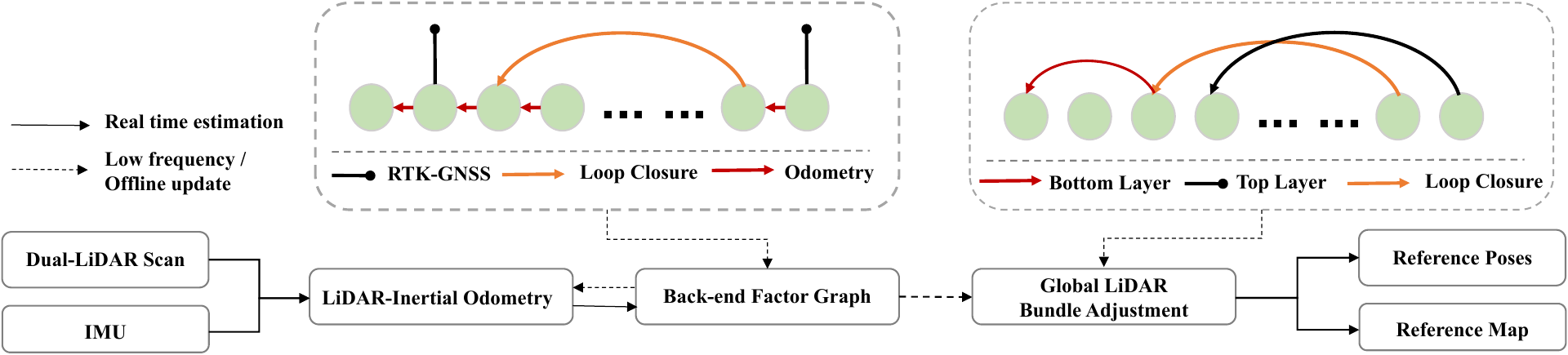}
\caption{Ground-Truth Generation Pipeline. A SLAM system with front-end/back-end interaction provides dense pose estimates, which are then globally corrected using RTK-GNSS data and further refined via offline global bundle adjustment.}
\label{fig_2}
\end{figure*}

LiDAR–IMU extrinsic calibration is essential to transform LiDAR data into the IMU(body) frame for point cloud de-skewing. Following the method \cite{lidar-imu-init}, we estimate the LiDAR–IMU extrinsic transform using smooth trajectories.

GNSS–IMU extrinsic calibration estimates the fixed spatial offset—known as the lever arm—between the GNSS antenna and the IMU. The GNSS receiver provides high-precision WGS84 coordinates $(\phi, \lambda, h)$ via RTK, timestamped using a PPS signal. Since SLAM typically adopts the initial IMU frame as the global reference, GNSS positions must be transformed accordingly to enable drift-free global constraints. To align both systems in a common Euclidean frame, we define an ENU (East-North-Up) coordinate system anchored at the initial GNSS position. Each measurement is converted from geodetic coordinates to a 3D ENU position vector $\mathbf{p}_i^{\text{GNSS}} \in \mathbb{R}^3$, consistent with the IMU and LiDAR outputs. We then leverage a smooth trajectory segment to construct residuals between the transformed GNSS positions and SLAM-estimated IMU poses. The lever arm vector $\mathbf{l}$ is obtained by minimizing the sum of squared residuals between the two.

Let $\mathbf{p}_i^{\text{GNSS}}$ denote the GNSS position at the $i$-th frame, $\mathbf{R}_i^{\text{IMU}}$ be the rotation matrix from the IMU frame to the world frame at the same time, and $\mathbf{l}$ be the GNSS-to-IMU lever arm vector to be estimated. The objective is to minimize the following cost function:

\begin{equation}
\mathcal{L} = \sum_{i=1}^{N} \left\| \mathbf{p}_i^{\text{GNSS}} - \left(\mathbf{p}_i^{\text{IMU}} + \mathbf{R}_i^{\text{IMU}} \cdot \mathbf{l} \right) \right\|^2
\end{equation}

where $\mathbf{p}_i^{\text{IMU}}$ is the IMU position estimated by SLAM. Minimizing $\mathcal{L}$ yields an optimal estimate of $\mathbf{l}$, completing the spatial alignment of GNSS measurements with the IMU frame.

Although the lever arm $\mathbf{l}$ can be estimated offline, the rotation between the IMU frame and the ENU frame varies randomly for each data collection session. Thus, we estimate this rotation online. We use the first 40 GNSS measurements, which are sampled at 0.5 m spatial intervals at the beginning of a sequence, to calibrate the rotation. The rotation calibration cost function is defined as:

\begin{equation}
\mathcal{L} = \sum_{i=1}^{N} \left\| \mathbf{R}_{\text{ENU}\rightarrow\text{IMU}} \cdot \mathbf{p}_i^{\text{GNSS}} + \mathbf{l} - \mathbf{p}_i^{\text{IMU}} \right\|^2
\end{equation}

Here, $\mathbf{R}_{\text{ENU}\rightarrow\text{IMU}}$ denotes the rotation matrix from the ENU frame to the IMU frame, $\mathbf{p}_i^{\text{GNSS}}$ is the GNSS position expressed in the ENU coordinate system, and $\mathbf{p}_i^{\text{IMU}}$ is the corresponding IMU position estimated by the LIO system.

\begin{figure}[!t]
\centering
\includegraphics[width=9cm]{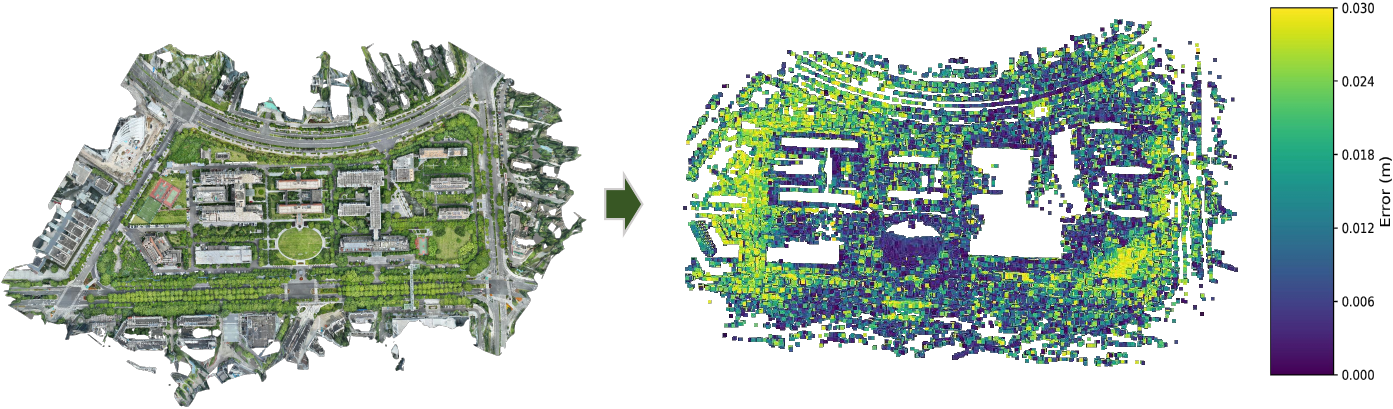}
\caption{Map accuracy visualization. The residuals are computed by aligning the generated ground-truth map with the UAV photogrammetric map, showing the error distribution.}
\label{residuald}
\end{figure}
\subsection{Reference Trajectory Generation}
As illustrated in Figure \ref{fig_2}, our ground-truth generation pipeline consists of three key steps. First, we build an initial high-precision map using data from the dual LiDARs combined with a state-of-the-art voxel-based SLAM algorithm \cite{Voxel-SLAM}. In the SLAM back-end, we fuse pre-calibrated RTK-GNSS measurements with the odometry estimates to obtain a drift-free global trajectory. Although RTK-GNSS provides accurate absolute position constraints, it lacks orientation information and is thus used only for real-time position correction during online SLAM. In the offline map optimization stage, we omit the RTK observations and instead refine the trajectory using loop closure detection and LiDAR-based bundle adjustment (BA).

Note that LiDAR BA optimizes the trajectory based on the map’s internal geometric consistency and cannot directly correct globally unobservable errors such as absolute height offsets or overall global drift. Therefore, we first use RTK-GNSS to anchor the initial map, eliminating any absolute position error. We then apply BA on top of this anchored map to suppress local trajectory noise and accumulated rotational drift, ultimately producing
continuous, dense, and low-drift 6-DoF ground-truth trajectories.

To validate the accuracy of our ground-truth trajectories, we conduct aerial oblique image capture in each scene using a DJI Mavic 3 drone equipped with RTK-GNSS. We generate centimeter-level global photogrammetric maps via DJI Terra and align them with our ground LiDAR maps for residual evaluation \cite{map_eval}. Figure \ref{residuald} shows that the alignment residuals between the LiDAR and aerial maps remain within a few centimeters throughout the area, demonstrating the robustness and accuracy of our method even in complex large-scale environments.

\begin{table*}[htp]
\caption{Comparison of Sequence Attributes in Our Dataset}
\label{Attributes}
\centering
\resizebox{14cm}{!}{
\begin{tabular}{ccccccccc}
\hline
\textbf{Features/Scene} & \multicolumn{3}{c}{\textbf{Park}} & \multicolumn{2}{c}{\textbf{Mountain}} & \textbf{Underground corridor} & \multicolumn{2}{c}{\textbf{Building complex}} \\ \hline
Sequences              & Seq.1      & Seq.2    & Seq.3     & Seq.4              & Seq.5            & Seq.6                         & Seq.7                 & Seq.8                 \\
Length /m              & 1561.82    & 1279     & 1002.05   & 1764.34            & 814.43                  & 1833.16                       & 1505.36               & 1307.83               \\
duration /s            & 1558.3     & 1174.9   & 1626.6    & 1688.1             &  1101.6                & 1791.8                        & 1849.6                & 1245.3                \\
Size /GB               & 82.0           & 65.3     & 88.7      & 97.2               & 62               & 87.5                          & 100.4                 & 70.9                  \\ \hline
\end{tabular}}
\end{table*}

\section{EXPERIMENT}
\subsection{Test Sequences}
To systematically evaluate LIO performance across diverse mapping scenarios, we collaborated with local surveying agencies to collect data in four representative environments. High-precision maps and continuous 6-DoF ground-truth trajectories were generated using our pipeline.

Table \ref{Attributes}shows first scene that including high-rise and underground structures, with three sequences traversing through tall buildings, basement parking areas, and long hallways. In this high-rise setting, RTK-GNSS signals were additionally obtained on a rooftop balcony to mitigate the loss of global constraints over the long trajectory. The second environment features unstructured mountainous terrain as well as the
interior of a tower. We recorded two sequences here: one over a steep hillside with sparse vegetation, and another inside a multi-story tower (including narrow staircases and floor transitions). And the third scene captures an underground utility tunnel. Operating in this confined space, the dual-LiDAR setup ensured full field of view coverage, enabling offline global BA to effectively eliminate accumulated drift. The final site is a dense cluster of buildings. We collected two sequences here: one around the perimeter and another weaving through complex indoor corridors, all yielding continuous trajectories spanning both indoor and outdoor spaces. 

These sequences span unstructured terrain, narrow environments, dense architecture, and extended indoor–outdoor transitions, offering comprehensive coverage for evaluating the robustness and generalization of LIO systems.

\begin{table*}[]
\caption{(Average / Maximum) Position Errors (in meters) of Different Algorithms on Each Sequence}
\label{position error}
\centering
\resizebox{18cm}{!}{
\begin{threeparttable}

\begin{tabular}{ccccccccc}
\hline
\textbf{Method/Scene} & \multicolumn{3}{c}{\textbf{Park}}                            & \multicolumn{2}{c}{\textbf{Mountain}} & \textbf{\begin{tabular}[c]{@{}c@{}}Underground corridor\end{tabular}} & \multicolumn{2}{c}{\textbf{Building complex}} \\ \hline
                      & Seq.1              & Seq.2              & Seq.3              & Seq.4                    & Seq.5      & Seq.6                                                                    & Seq.7                 & Seq.8                 \\
d\_lio\cite{DLIO}                & 6.27/11.36         & 1.79/4.11          & 12.48/78.00        & 6.76/15.19               & *          & *                                                                        & 1.23/3.27             & 3.96/9.18             \\
fast\_lio2\cite{FAST-LIO2}            & 9.25/21.87         & 3.26/11.36         & *                  & 9.63/40.94               & *          & *                                                                        & 1.65/5.57             & 4.01/10.45            \\
ig\_lio\cite{ig-lio}               & 7.52/16.84         & 2.86/9.77          & 9.896/24.218       & *                        & *          & *                                                                        & 3.69/11.59            & 4.90/13.23            \\
point\_lio\cite{Point-LIO}            & 3.84/8.02          & 4.63/10.15         & *                  & 16.52/51.37              & *          & *                                                                        & 0.44/2.25             & 2.89/5.96             \\
lio\_sam\cite{LIO-SAM}              & 11.75/20.69        & 2.47/6.42          & *                  & 21.38/74.64              & *          & *                                                                        & 2.06/7.26             & 4.07/10.58            \\
voxel\_slam\cite{Voxel-SLAM}           & \textbf{0.34/1.07} & \textbf{0.72/2.03} & \textbf{0.35/0.72} & \textbf{2.02/3.99}       & *          & \textbf{7.95/*}                                                          & \textbf{0.04/0.12}    & \textbf{0.42/0.93}    \\ \hline
\end{tabular}

 \begin{tablenotes}
        \footnotesize
        \item[*] Indicates degraded results where the position error exceeds 100 meters or the trajectory fails to complete
      \end{tablenotes}
    \end{threeparttable}}

\end{table*}

\subsection{Evaluation Setup}
To evaluate LIO performance in long-duration, continuous indoor–outdoor mapping scenarios, we selected several representative methods, each chosen for its suitability to specific degraded conditions. IG-LIO \cite{ig-lio} uses adaptive geometric consistency, suitable for corridors and repetitive structures. FAST-LIO2 \cite{FAST-LIO2} applies iterative Kalman filtering with direct scan-to-map matching, robust under high-speed motion. D-LIO \cite{DLIO} leverages continuous-time trajectories and coarse-to-fine matching, designed for resource-constrained or dynamic settings. Point-LIO \cite{Point-LIO} models with features distributions per point perform well in sparse vegetation and open areas. LIO-SAM \cite{LIO-SAM} introduces loop closure and global optimization, ideal for environments with frequent revisits. Voxel-SLAM \cite{Voxel-SLAM} combines voxel mapping with loop optimization to maintain global consistency in large-scale, seamless indoor–outdoor environments. Together, these selected methods cover both filter-based and graph-based LIO frameworks, including approaches with and without loop closure support. All algorithms process data from a single LiDAR and IMU for pose estimation.

We evaluate each algorithm using three metrics: (1) the mean and maximum per frame Euclidean position error; (2) the mean and maximum orientation error per frame (in degrees); and (3) the peak RAM usage (in GB) during the full sequence, as a measure of runtime resource demand.

\begin{figure}[!t]
\centering
\subfigure[Seq.1]{\includegraphics[width=4.2cm]{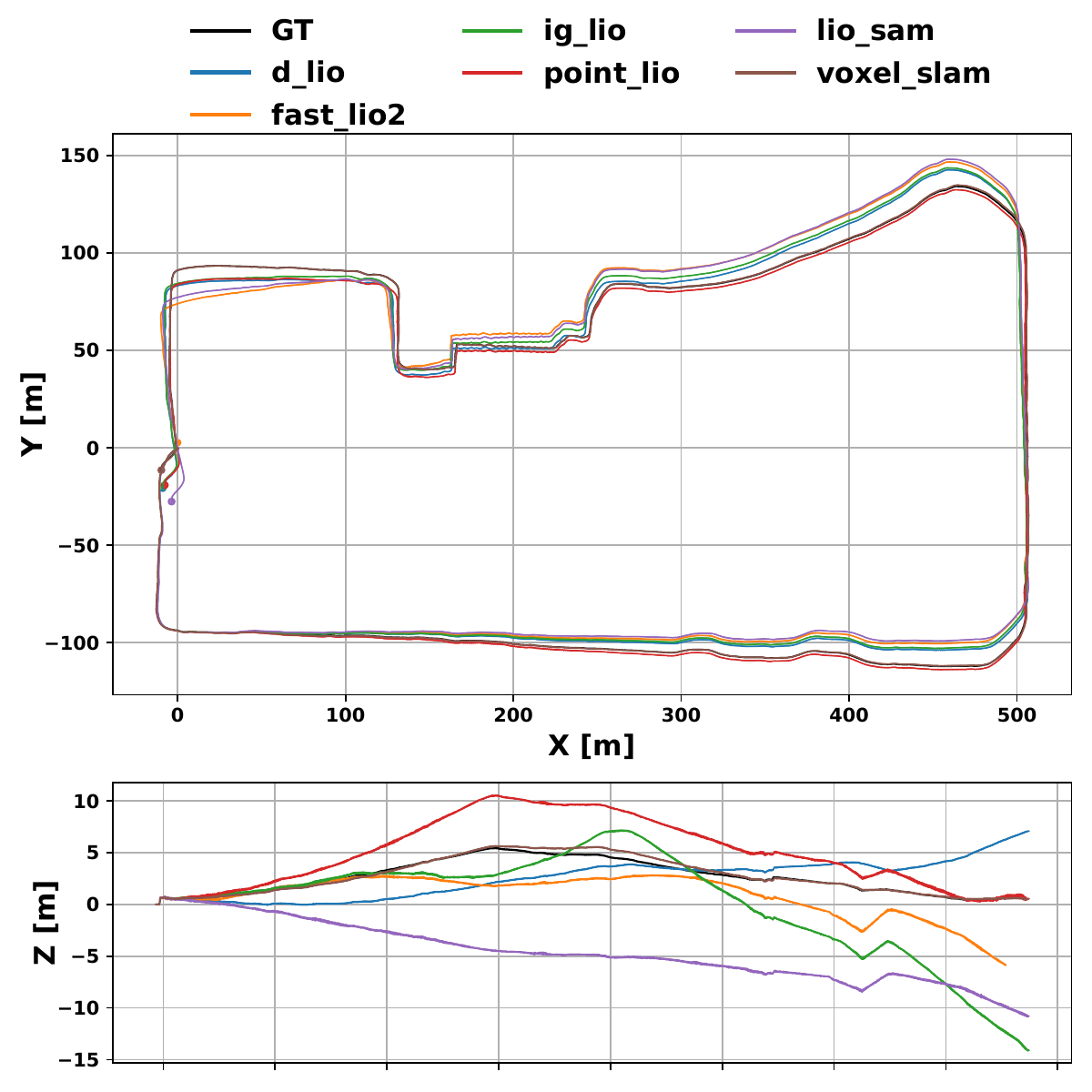}}\subfigure[Seq.2]{\includegraphics[width=4.2cm]{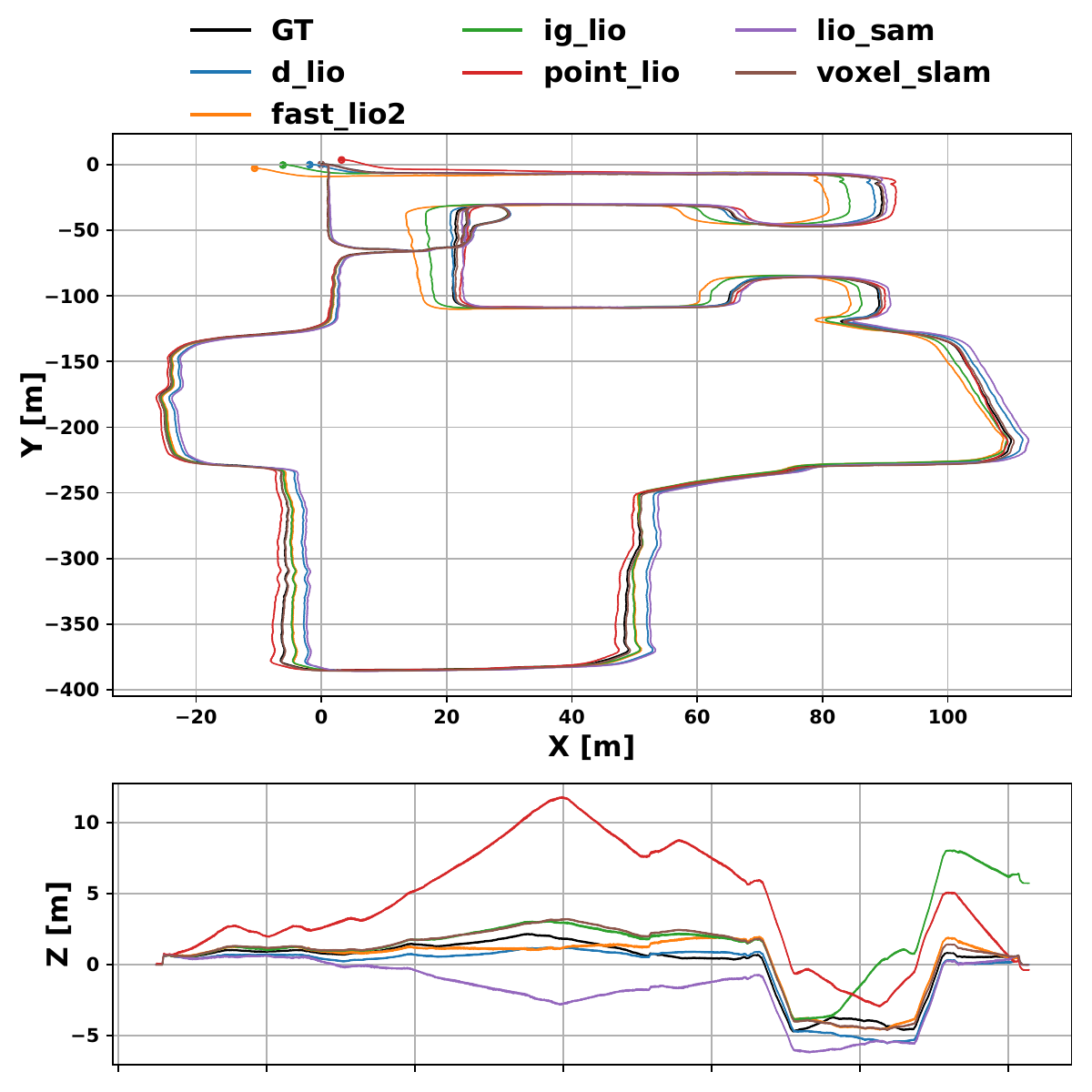}}
\subfigure[Seq.3]{\includegraphics[width=4.2cm]{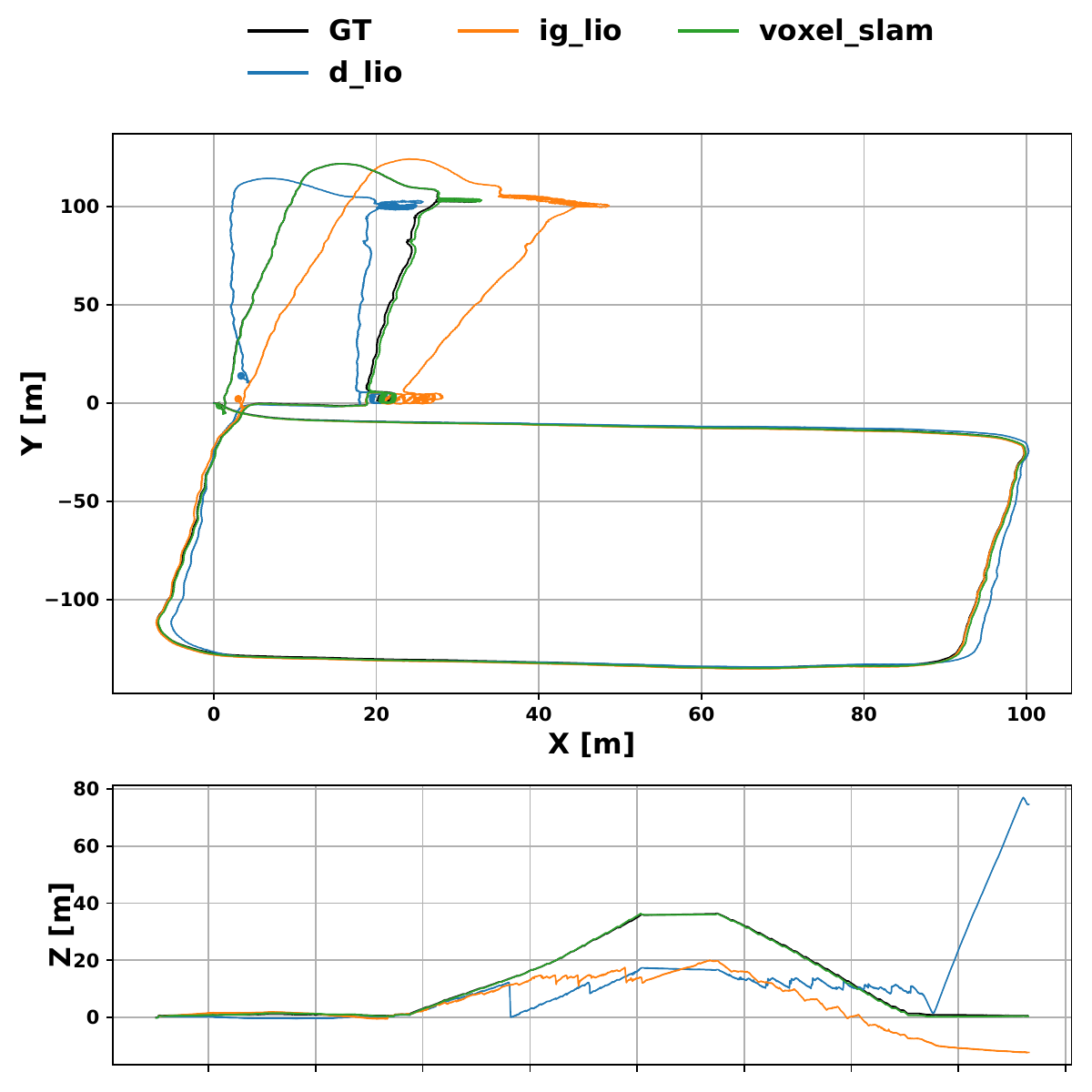}}\subfigure[Seq.4]{\includegraphics[width=4.2cm]{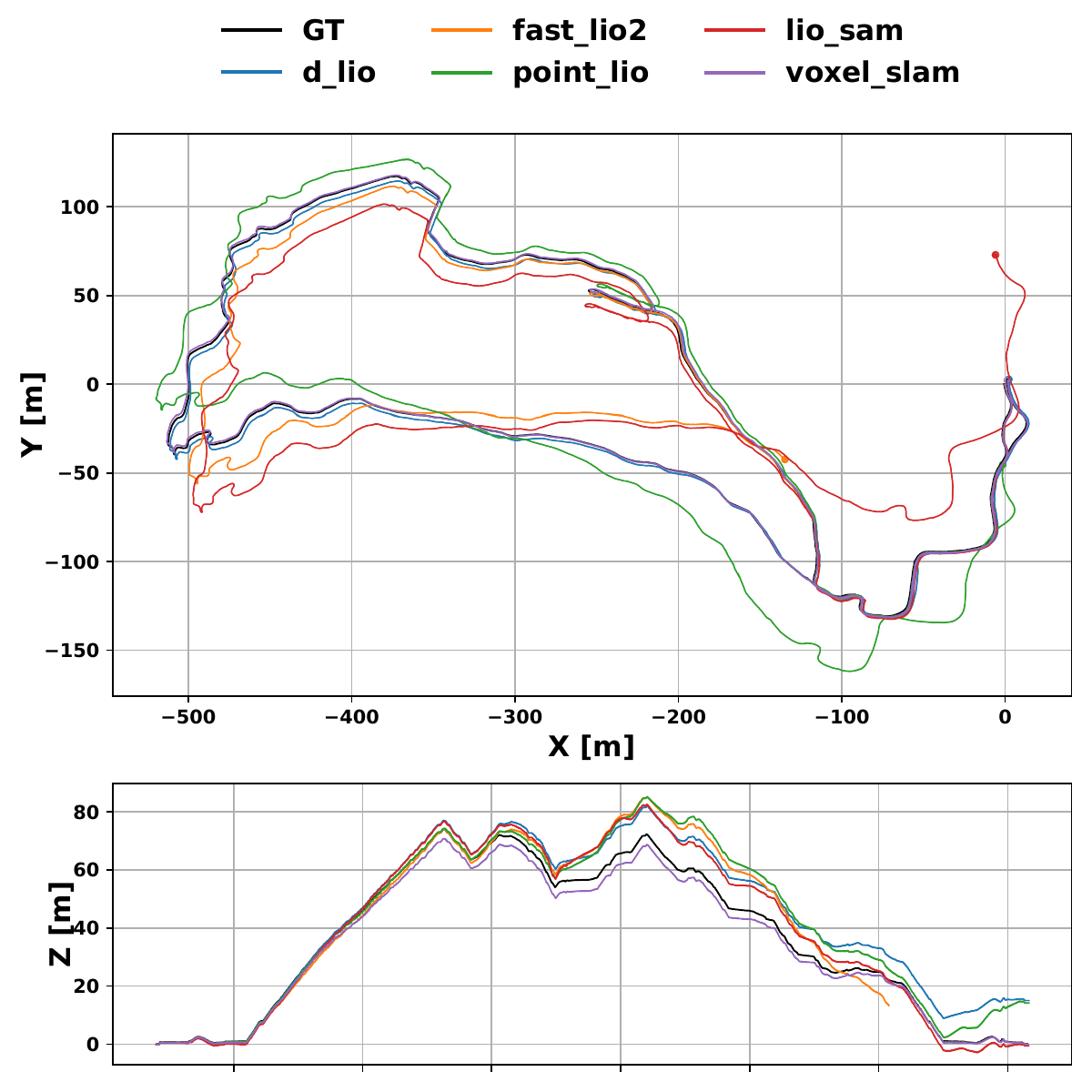}}
\subfigure[Seq.7]{\includegraphics[width=4.2cm]{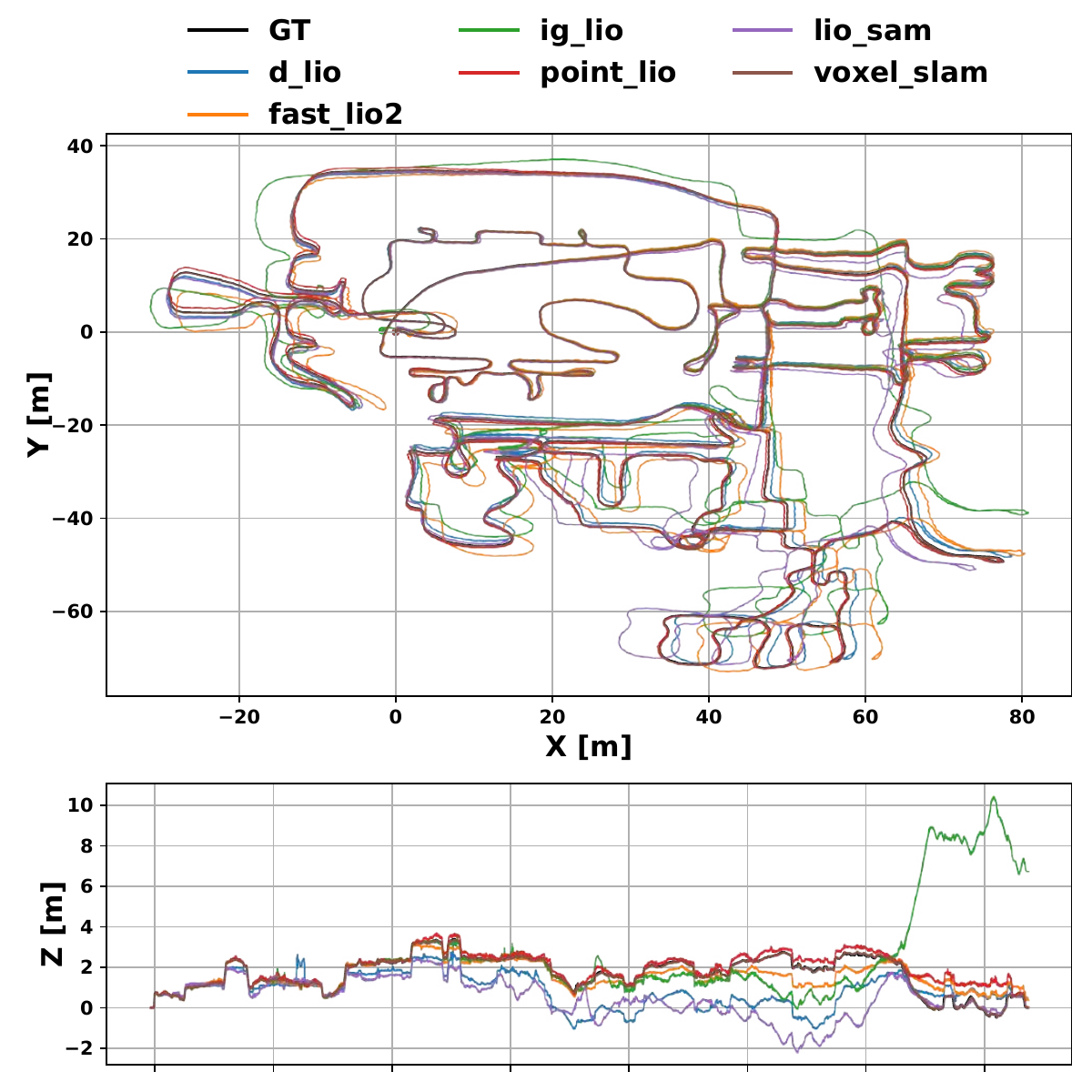}}\subfigure[Seq.8]{\includegraphics[width=4.2cm]{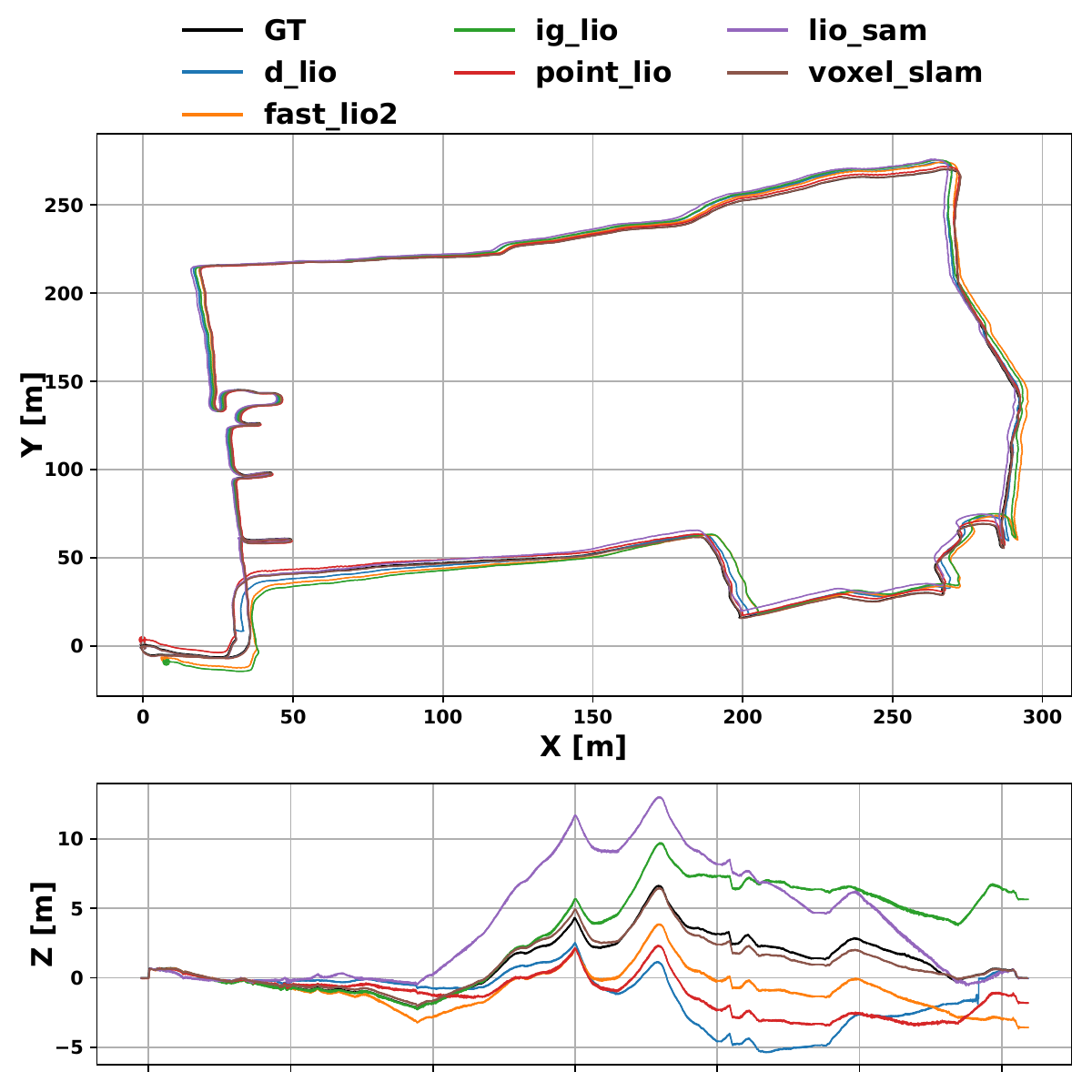}}
\caption{Trajectory comparison. Estimated trajectories from each method over multiple test sequences, facilitating comparative analysis of localization performance.}
\label{fig_8}
\end{figure}

\subsection{Evaluation Results}

\subsubsection{Pose Accuracy}
As shown in Figure \ref{fig_8}, most algorithms exhibit noticeable vertical drift in large-scale mapping sequences, particularly in open or weakly structured environments. This consistent altitude deviation reflects a fundamental limitation of current LIO pipelines: the lack of strong constraints along the gravity axis in the absence of GNSS or geometric priors. Even advanced systems with loop closure or adaptive filtering struggle to suppress accumulated vertical errors over long trajectories.

From the quantitative results in Table \ref{position error} and \ref{angle error}, Voxel-SLAM achieves the lowest average pose error across sequences. This can be attributed to its maintenance of global map consistency through tightly coupled front- and back-end components. Unlike methods that rely solely on local submaps, Voxel-SLAM continuously enforces long-range constraints via a unified voxel representation. However, the accuracy achieved still falls short of the centimeter-level precision required for professional mapping applications, indicating room for further improvement in both trajectory optimization and global consistency under challenging conditions.

Regarding loop closure, LIO-SAM relies on Euclidean proximity for loop detection, which often fails in large-scale environments where revisited locations may be far apart or geometrically ambiguous. In contrast, Voxel-SLAM leverages a compact descriptor for place recognition, enabling robust loop closure even across scenes with large drift or perceptual aliasing. This capability is crucial for maintaining global consistency in long-term, drift-prone mapping scenarios.

\subsubsection{Robutness}

Under the evaluated degenerate conditions, D-LIO provided the most consistently reliable odometry, remaining effective across most environments. Voxel-SLAM achieved the best overall performance. Its multi-session design allows the system to start a new submap after severe drift and later reconnect it to previous maps via place recognition, ensuring global consistency under long-term degradation.

\subsubsection{Resources Cost}
As shown in Table \ref{ram usage}, the memory consumption of the different
algorithms varies significantly across the sequences. Voxel-SLAM, which maintains a global voxel map, shows steadily increasing RAM usage as scene size grows — reaching nearly 15 GB in the largest environment — indicating potential issues for long-term mapping on memory-limited platforms. In contrast, D-LIO’s lightweight design keeps memory usage around 1 GB. Combined
with its competitive accuracy, this low resource footprint makes D-LIO a strong candidate for resource-constrained applications.
\begin{figure*}[htp]
\centering
\subfigure[Seq.3]{\includegraphics[width=5.8cm]{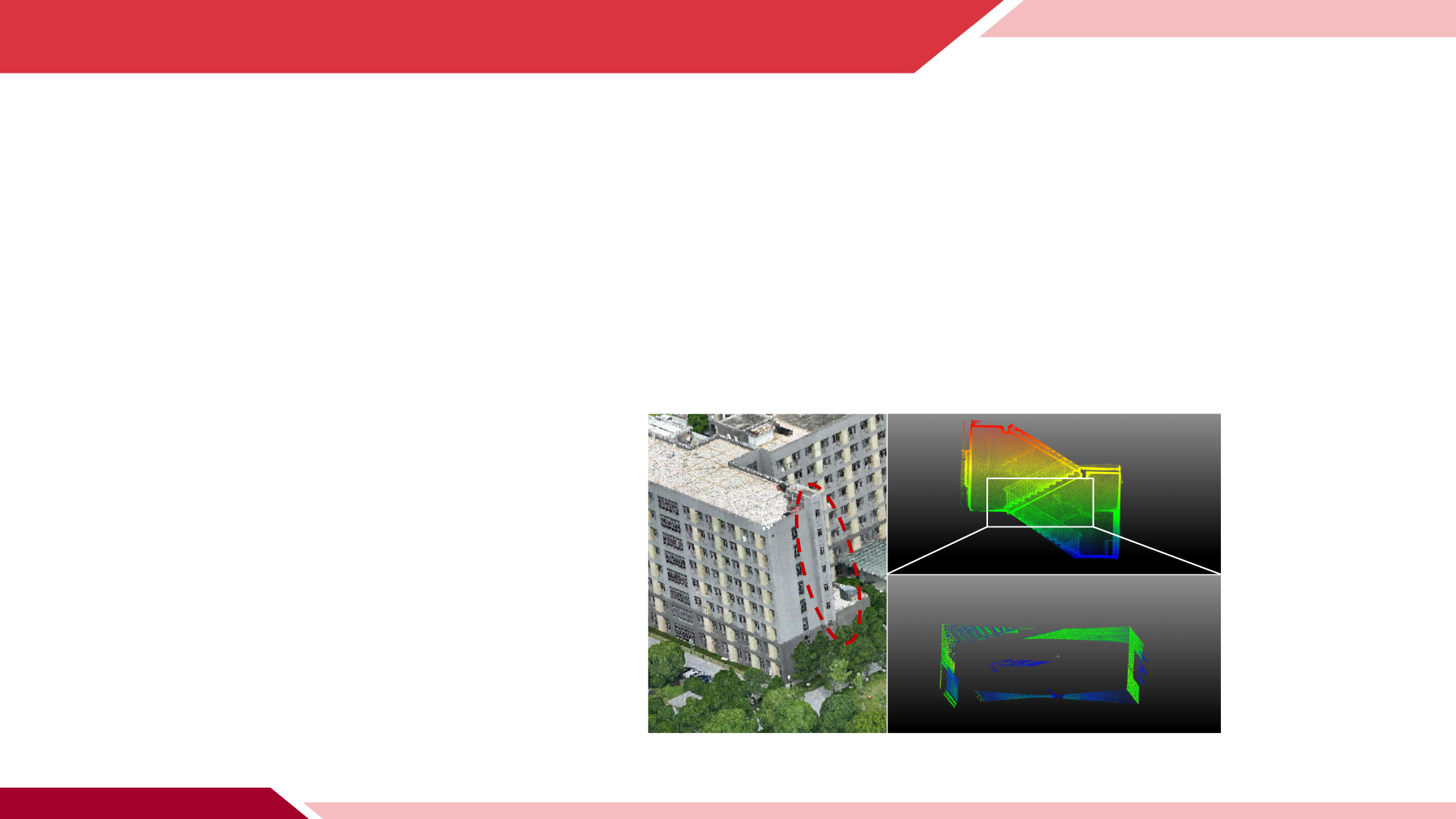}}
\subfigure[Seq.5]{\includegraphics[width=5.8cm]{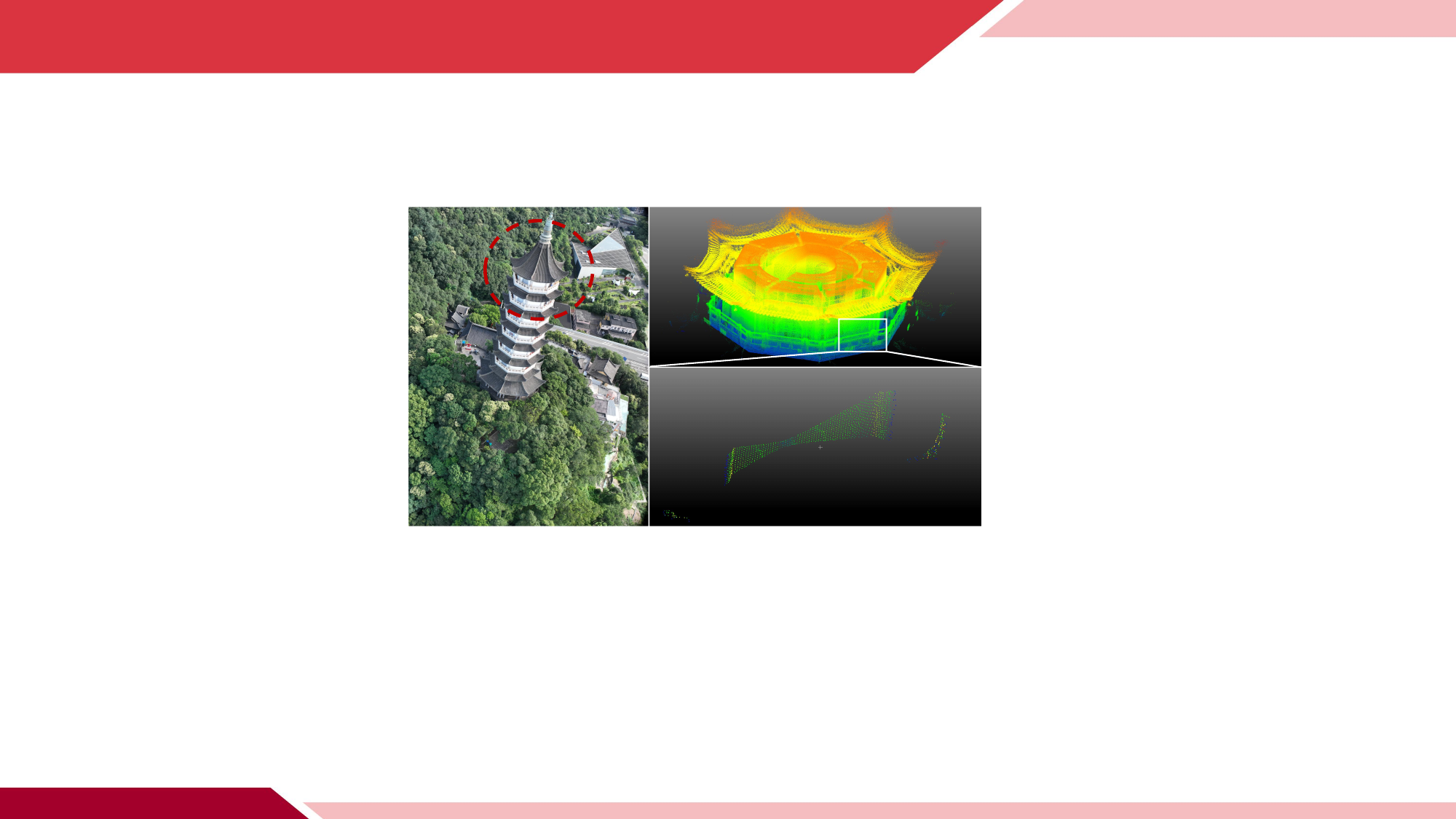}}
\subfigure[Seq.6]{\includegraphics[width=5.8cm]{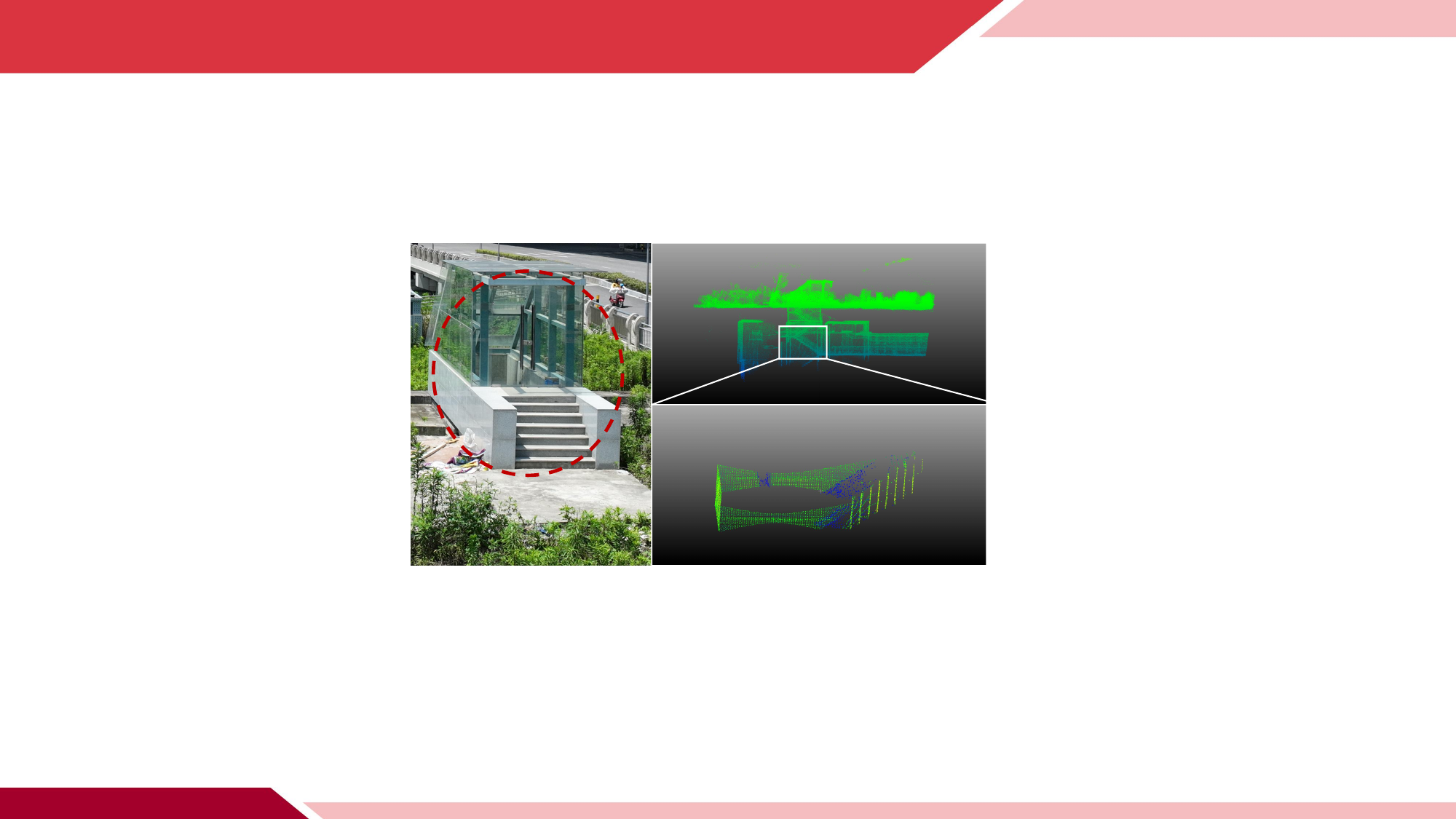}}
\caption{Analysis of degeneration. The sequences where LIO algorithms encounter degeneration, with highlighted LiDAR frames showing under-constrained regions.}
\label{degration}
\end{figure*}

\begin{table*}[]
\caption{(Average / Maximum) Angle Errors (in degrees) of Different Algorithms on Each Sequence}
\label{angle error}
\centering
\resizebox{16cm}{!}{
\begin{threeparttable}
\begin{tabular}{ccccccccc}
\hline
\textbf{Method/Scene} & \multicolumn{3}{c}{\textbf{Park}}                            & \multicolumn{2}{c}{\textbf{Mountain}} & \textbf{Underground corridor} & \multicolumn{2}{c}{\textbf{Building complex}} \\ \hline
                      & Seq.1              & Seq.2              & Seq.3              & Seq.4                    & Seq.5      & Seq.6                         & Seq.7                 & Seq.8                 \\
d\_lio \cite{DLIO}               & 2.20/6.19          & 1.09/5.20          & 14.78/*            & 1.96/6.05                & *          & *                             & 1.83/8.87             & 1.86/8.08             \\
fast\_lio2 \cite{FAST-LIO2}            & 3.23/10.79         & 1.82/6.80          & *                  & 4.61/13.66               & *          & *                             & 3.23/10.96            & 1.62/5.79             \\
ig\_lio \cite{ig-lio}               & 2.69/7.50          & 1.66/5.92          & 15.29/15.29        & *                        & *          & *                             & 4.97/14.81            & 2.20/8.07             \\
point\_lio \cite{Point-LIO}            & 1.16/6.38          & 2.53/4.70          & *                  & 5.55/13.98               & *          & *                             & 1.23/7.63             & 1.20/5.20             \\
lio\_sam \cite{LIO-SAM}              & 3.15/9.27          & 1.19/4.96          & *                  & 7.05/17.22               & *          & *                             & 3.86/14.14            & 1.67/5.41             \\
voxel\_slam \cite{Voxel-SLAM}           & \textbf{0.17/0.91} & \textbf{0.72/1.49} & \textbf{0.69/3.00} & \textbf{0.70/2.68}       & *          & \textbf{3.03/*}               & \textbf{0.28/1.92}    & \textbf{0.43/2.16}    \\ \hline
\end{tabular}

 \begin{tablenotes}
        \footnotesize
        \item[*] Indicates degraded results where the position error exceeds 20 degrees or the trajectory fails to complete
      \end{tablenotes}
    \end{threeparttable}}

\end{table*}

\begin{table*}[]
\caption{Peak RAM Usage (in GB) of Different Algorithms During Execution}
\label{ram usage}
\centering
\resizebox{16cm}{!}{
\begin{threeparttable}
\begin{tabular}{ccccccccc}
\hline
\textbf{Method/Scene} & \multicolumn{3}{c}{\textbf{Park}}                   & \multicolumn{2}{c}{\textbf{Mountain}} & \textbf{Underground corridor} & \multicolumn{2}{c}{\textbf{Building complex}} \\ \hline
                      & Seq.1           & Seq.2           & Seq.3           & Seq.4                  & Seq.5        & Seq.6                         & Seq.7                  & Seq.8                \\
d\_lio \cite{DLIO}                & \textbf{1.2361} & 1.008           & \textbf{0.6323} & 1.3877                 & *            & *                             & 0.945                  & 0.9351               \\
fast\_lio2 \cite{FAST-LIO2}            & 14.816          & 7.8646          & *               & 15.323                 & *            & *                             & 2.909                  & 16.2878              \\
ig\_lio \cite{ig-lio}               & 2.6019          & 1.7442          & 1.1768          & 2.9683                 & *            & *                             & 0.7239                 & 1.6806               \\
point\_lio \cite{Point-LIO}            & 2.9683          & \textbf{0.7435} & *               & \textbf{0.8965}        & *            & *                             & \textbf{0.6313}        & \textbf{0.703}       \\
lio\_sam \cite{LIO-SAM}              & 6.5825          & 5.7607          & *               & 5.8103                 & *            & *                             & 5.0636                 & 5.1157               \\
voxel\_slam \cite{Voxel-SLAM}           & 16.7032         & 15.1806         & 11.9088         & 10.5003                & *            & 13.9458                    & 5.1275                 & 13.2183              \\ \hline
\end{tabular}

 \begin{tablenotes}
        \footnotesize
        \item[*] Indicates degraded results where the trajectory fails to complete
      \end{tablenotes}
    \end{threeparttable}}

\end{table*}

\subsubsection{Degradation Analysis}
Figure \ref{degration} highlights typical degrade scenarios encountered in sequences 3, 5, and 6. In Sequence 3, the main challenge is the transition from an open outdoor space to a narrow staircase, which makes it difficult for the algorithms to utilize the sparse geometric features effectively. Sequence 5 features a tower structure where the outer corridors offer a minimal field of view and weak geometric constraints. In Sequence 6, a degrade case occurs near a staircase at the end of a long corridor, again due to very sparse structural features. All experiments were conducted using each algorithm’s indoor configuration settings. If we used the algorithms’ outdoor settings instead, then handling transitions into constrained spaces would require dynamic adaptation (such as adjusting sampling density) to prevent data loss and reduce degeneration.

\subsection{Discussion}
Due to their reliance on predefined parameters, most LIO algorithms are typically configured separately for indoor (narrow) and outdoor (open) environments. This fixed-parameter paradigm limits adaptability to diverse scenes and compromises the ability to achieve optimal performance and estimation accuracy. Moreover, LIO systems often exhibit limited robustness to local degeneracies, where degraded segments hinder subsequent scan registration. Even in multi-session SLAM frameworks, frequent loop closures in the back end are required to mitigate front-end failures, ultimately reducing overall generalizability. These limitations underscore the need for adaptive LIO algorithms capable of dynamically adjusting to environmental variations to ensure high-precision and robust pose estimation across diverse situation.

\section{Conclusion}
In this work, we presented a new large-scale LiDAR–IMU benchmark dataset specifically designed for high-precision mapping tasks in real-world environments, featuring long sequences and seamless indoor–outdoor transitions. The dataset was collected using a wearable multi-sensor platform and processed through a high-accuracy GNSS-assisted LiDAR mapping pipeline, providing centimeter-level 6-DoF ground truth even under various degraded conditions. We conducted comprehensive evaluations of several representative LIO algorithms, revealing their respective strengths and limitations in challenging scenarios such as sparse features, repetitive structures, GNSS-denied areas, and long-term drift. Overall, the proposed benchmark fills an important gap in the current set of available datasets and offers a standardized platform for testing more robust, generalizable LIO systems in practical applications.

\section*{Acknowledgments}
This work was supported in part by the Key R\&D Program of Zhejiang 2025C01075, in part by the National Natural Science Foundation of China under Grant U24A20249, in part by the Key R\&D Program of Ningbo under Grant 2023Z220, 2023Z224, 2024Z300, 2025Z061, in part by the Open Fund of the Technology Innovation Center for 3D Real Scene Construction and Urban Refined Governance,Ministry of Natural Resources (Grant No.2024PF-3), and in part by the Public Welfare Science and Technology Plan Project of Ningbo under Grant 2024S061.


\vfill


\begin{thebibliography}{10}
\providecommand{\url}[1]{#1}
\csname url@samestyle\endcsname
\providecommand{\newblock}{\relax}
\providecommand{\bibinfo}[2]{#2}
\providecommand{\BIBentrySTDinterwordspacing}{\spaceskip=0pt\relax}
\providecommand{\BIBentryALTinterwordstretchfactor}{4}
\providecommand{\BIBentryALTinterwordspacing}{\spaceskip=\fontdimen2\font plus
\BIBentryALTinterwordstretchfactor\fontdimen3\font minus
  \fontdimen4\font\relax}
\providecommand{\BIBforeignlanguage}[2]{{%
\expandafter\ifx\csname l@#1\endcsname\relax
\typeout{** WARNING: IEEEtran.bst: No hyphenation pattern has been}%
\typeout{** loaded for the language `#1'. Using the pattern for}%
\typeout{** the default language instead.}%
\else
\language=\csname l@#1\endcsname
\fi
#2}}
\providecommand{\BIBdecl}{\relax}
\BIBdecl

\bibitem{NewerCollege}
M.~Ramezani, Y.~Wang, M.~Camurri, D.~Wisth, M.~Mattamala, and M.~Fallon, ``The newer college dataset: Handheld lidar, inertial and vision with ground truth,'' in \emph{2020 IEEE/RSJ International Conference on Intelligent Robots and Systems (IROS)}.\hskip 1em plus 0.5em minus 0.4em\relax IEEE, 2020, pp. 4353--4360.

\bibitem{HiltiSLAM}
M.~Helmberger, K.~Morin, B.~Berner, N.~Kumar, G.~Cioffi, and D.~Scaramuzza, ``The hilti slam challenge dataset,'' \emph{IEEE Robotics and Automation Letters}, vol.~7, no.~3, pp. 7518--7525, 2022.

\bibitem{Seasons}
P.~Wenzel, R.~Wang, N.~Yang, Q.~Cheng, Q.~Khan, L.~Von~Stumberg, N.~Zeller, and D.~Cremers, ``4seasons: A cross-season dataset for multi-weather slam in autonomous driving,'' in \emph{DAGM German Conference on Pattern Recognition}.\hskip 1em plus 0.5em minus 0.4em\relax Springer, 2020, pp. 404--417.

\bibitem{ComplexUrban}
J.~Jeong, Y.~Cho, Y.-S. Shin, H.~Roh, and A.~Kim, ``Complex urban lidar data set,'' in \emph{2018 IEEE international conference on robotics and automation (ICRA)}.\hskip 1em plus 0.5em minus 0.4em\relax IEEE, 2018, pp. 6344--6351.

\bibitem{Montmorency}
J.-F. Tremblay, M.~B{\'e}land, R.~Gagnon, F.~Pomerleau, and P.~Gigu{\`e}re, ``Automatic three-dimensional mapping for tree diameter measurements in inventory operations,'' \emph{Journal of Field Robotics}, vol.~37, no.~8, pp. 1328--1346, 2020.

\bibitem{LOAM}
J.~Zhang, S.~Singh \emph{et~al.}, ``Loam: Lidar odometry and mapping in real-time.'' in \emph{Robotics: Science and systems}, vol.~2, no.~9.\hskip 1em plus 0.5em minus 0.4em\relax Berkeley, CA, 2014, pp. 1--9.

\bibitem{LIO-SAM}
T.~Shan, B.~Englot, D.~Meyers, W.~Wang, C.~Ratti, and D.~Rus, ``Lio-sam: Tightly-coupled lidar inertial odometry via smoothing and mapping,'' in \emph{2020 IEEE/RSJ international conference on intelligent robots and systems (IROS)}.\hskip 1em plus 0.5em minus 0.4em\relax IEEE, 2020, pp. 5135--5142.

\bibitem{BEV-LIO}
H.~Cai, S.~Yuan, X.~Li, J.~Guo, and J.~Liu, ``Bev-lio (lc): Bev image assisted lidar-inertial odometry with loop closure,'' \emph{arXiv preprint arXiv:2502.19242}, 2025.

\bibitem{IPAL}
X.~Jin, J.~Ge, J.~Xiao, N.~Bu, and G.~Xu, ``Ipal: Infinite planes as lines for consistent mapping in indoor multifloor environments,'' \emph{IEEE Transactions on Instrumentation and Measurement}, vol.~73, pp. 1--11, 2024.

\bibitem{FAST-LIO2}
W.~Xu, Y.~Cai, D.~He, J.~Lin, and F.~Zhang, ``Fast-lio2: Fast direct lidar-inertial odometry,'' \emph{IEEE Transactions on Robotics}, vol.~38, no.~4, pp. 2053--2073, 2022.

\bibitem{faster-LIO}
C.~Bai, T.~Xiao, Y.~Chen, H.~Wang, F.~Zhang, and X.~Gao, ``Faster-lio: Lightweight tightly coupled lidar-inertial odometry using parallel sparse incremental voxels,'' \emph{IEEE Robotics and Automation Letters}, vol.~7, no.~2, pp. 4861--4868, 2022.

\bibitem{adalio}
H.~Lim, D.~Kim, B.~Kim, and H.~Myung, ``Adalio: Robust adaptive lidar-inertial odometry in degenerate indoor environments,'' in \emph{2023 20th International Conference on Ubiquitous Robots (UR)}.\hskip 1em plus 0.5em minus 0.4em\relax IEEE, 2023, pp. 48--53.

\bibitem{DLIO}
K.~Chen, R.~Nemiroff, and B.~T. Lopez, ``Direct lidar-inertial odometry: Lightweight lio with continuous-time motion correction,'' \emph{arXiv preprint arXiv:2203.03749}, 2022.

\bibitem{COIN-LIO}
P.~Pfreundschuh, H.~Oleynikova, C.~Cadena, R.~Siegwart, and O.~Andersson, ``Coin-lio: Complementary intensity-augmented lidar inertial odometry,'' in \emph{2024 IEEE International Conference on Robotics and Automation (ICRA)}.\hskip 1em plus 0.5em minus 0.4em\relax IEEE, 2024, pp. 1730--1737.

\bibitem{NV-LIOM}
D.~Chung and J.~Kim, ``Nv-liom: Lidar-inertial odometry and mapping using normal vectors towards robust slam in multifloor environments,'' \emph{IEEE Robotics and Automation Letters}, 2024.

\bibitem{MM-LINS}
Y.~Ma, J.~Xu, S.~Yuan, T.~Zhi, W.~Yu, J.~Zhou, and L.~Xie, ``Mm-lins: a multi-map lidar-inertial system for over-degenerate environments,'' \emph{IEEE Transactions on Intelligent Vehicles}, 2024.

\bibitem{Voxel-SLAM}
Z.~Liu, H.~Li, C.~Yuan, X.~Liu, J.~Lin, R.~Li, C.~Zheng, B.~Zhou, W.~Liu, and F.~Zhang, ``Voxel-slam: A complete, accurate, and versatile lidar-inertial slam system,'' \emph{arXiv preprint arXiv:2410.08935}, 2024.

\bibitem{KITTI-360}
Y.~Liao, J.~Xie, and A.~Geiger, ``Kitti-360: A novel dataset and benchmarks for urban scene understanding in 2d and 3d,'' \emph{IEEE Transactions on Pattern Analysis and Machine Intelligence}, vol.~45, no.~3, pp. 3292--3310, 2022.

\bibitem{Hilti-Oxford}
L.~Zhang, M.~Helmberger, L.~F.~T. Fu, D.~Wisth, M.~Camurri, D.~Scaramuzza, and M.~Fallon, ``Hilti-oxford dataset: A millimeter-accurate benchmark for simultaneous localization and mapping,'' \emph{IEEE Robotics and Automation Letters}, vol.~8, no.~1, pp. 408--415, 2022.

\bibitem{M2DGR}
J.~Yin, A.~Li, T.~Li, W.~Yu, and D.~Zou, ``M2dgr: A multi-sensor and multi-scenario slam dataset for ground robots,'' \emph{IEEE Robotics and Automation Letters}, vol.~7, no.~2, pp. 2266--2273, 2021.

\bibitem{UrbanNav}
L.-T. Hsu, N.~Kubo, W.~Wen, W.~Chen, Z.~Liu, T.~Suzuki, and J.~Meguro, ``Urbannav: An open-sourced multisensory dataset for benchmarking positioning algorithms designed for urban areas,'' in \emph{Proceedings of the 34th international technical meeting of the satellite division of the institute of navigation (ION GNSS+ 2021)}, 2021, pp. 226--256.

\bibitem{WHU-Helmet}
J.~Li, W.~Wu, B.~Yang, X.~Zou, Y.~Yang, X.~Zhao, and Z.~Dong, ``Whu-helmet: A helmet-based multisensor slam dataset for the evaluation of real-time 3-d mapping in large-scale gnss-denied environments,'' \emph{IEEE Transactions on Geoscience and Remote Sensing}, vol.~61, pp. 1--16, 2023.

\bibitem{GEODE}
Z.~Chen, Y.~Qi, D.~Feng, X.~Zhuang, H.~Chen, X.~Hu, J.~Wu, K.~Peng, and P.~Lu, ``Heterogeneous lidar dataset for benchmarking robust localization in diverse degenerate scenarios,'' \emph{The International Journal of Robotics Research}, p. 02783649251344967, 2024.

\bibitem{GRACO}
Y.~Zhu, Y.~Kong, Y.~Jie, S.~Xu, and H.~Cheng, ``Graco: A multimodal dataset for ground and aerial cooperative localization and mapping,'' \emph{IEEE Robotics and Automation Letters}, vol.~8, no.~2, pp. 966--973, 2023.

\bibitem{FusionPortable}
J.~Jiao, H.~Wei, T.~Hu, X.~Hu, Y.~Zhu, Z.~He, J.~Wu, J.~Yu, X.~Xie, H.~Huang \emph{et~al.}, ``Fusionportable: A multi-sensor campus-scene dataset for evaluation of localization and mapping accuracy on diverse platforms,'' in \emph{2022 IEEE/RSJ International Conference on Intelligent Robots and Systems (IROS)}.\hskip 1em plus 0.5em minus 0.4em\relax IEEE, 2022, pp. 3851--3856.

\bibitem{MulRan}
G.~Kim, Y.~S. Park, Y.~Cho, J.~Jeong, and A.~Kim, ``Mulran: Multimodal range dataset for urban place recognition,'' in \emph{2020 IEEE international conference on robotics and automation (ICRA)}.\hskip 1em plus 0.5em minus 0.4em\relax IEEE, 2020, pp. 6246--6253.

\bibitem{BotanicGarden}
Y.~Liu, Y.~Fu, M.~Qin, Y.~Xu, B.~Xu, F.~Chen, B.~Goossens, P.~Z. Sun, H.~Yu, C.~Liu \emph{et~al.}, ``Botanicgarden: A high-quality dataset for robot navigation in unstructured natural environments,'' \emph{IEEE Robotics and Automation Letters}, vol.~9, no.~3, pp. 2798--2805, 2024.

\bibitem{lidar-imu-init}
F.~Zhu, Y.~Ren, and F.~Zhang, ``Robust real-time lidar-inertial initialization,'' in \emph{2022 IEEE/RSJ International Conference on Intelligent Robots and Systems (IROS)}.\hskip 1em plus 0.5em minus 0.4em\relax IEEE, 2022, pp. 3948--3955.

\bibitem{Point-LIO}
D.~He, W.~Xu, N.~Chen, F.~Kong, C.~Yuan, and F.~Zhang, ``Point-lio: robust high-bandwidth light detection and ranging inertial odometry,'' \emph{Advanced Intelligent Systems}, vol.~5, no.~7, p. 2200459, 2023.

\bibitem{ig-lio}
Z.~Chen, Y.~Xu, S.~Yuan, and L.~Xie, ``ig-lio: An incremental gicp-based tightly-coupled lidar-inertial odometry,'' \emph{IEEE Robotics and Automation Letters}, vol.~9, no.~2, pp. 1883--1890, 2024.

\bibitem{map_eval}
X.~Hu, J.~Wu, M.~Jia, H.~Yan, Y.~Jiang, B.~Jiang, W.~Zhang, W.~He, and P.~Tan, ``Mapeval: towards unified, robust and efficient slam map evaluation framework,'' \emph{IEEE Robotics and Automation Letters}, 2025.

\bibitem{lidar_cali}
H.~Ye, Y.~Jin, J.~Liu, T.~Li, W.~Zhang, and M.~Fu, ``Dlbacalib: Robust extrinsic calibration for non‑overlapping lidars based on dual lba,'' \emph{arXiv preprint arXiv:2507.09176}, 2025.
\end{thebibliography}
\end{document}